\documentclass[runningheads]{llncs}
\usepackage{graphicx}
\usepackage{amsmath,amssymb} 
\usepackage{color}
\usepackage{xcolor}
\usepackage{pifont}
\usepackage{caption}
\usepackage{subcaption}
\usepackage{booktabs}
\usepackage[moderate]{savetrees} 
\usepackage{hyperref}
\usepackage{cleveref}[2012/02/15]

\begin{document}
\pagestyle{headings}
\mainmatter

\def\ACCV20SubNumber{13}  

\title{\mbox{Condensed Movies}: \mbox{Story Based Retrieval with Contextual Embeddings}} 
\titlerunning{Condensed Movies}
\authorrunning{M. Bain et al.}

\author{Max Bain \and Arsha Nagrani \and Andrew Brown \and Andrew Zisserman}
\institute{Visual Geometry Group, Department of Engineering Science, University of Oxford\\
\email{\{maxbain,arsha,abrown,az\}@robots.ox.ac.uk}
} 

\maketitle
\begin{abstract}
Our objective in this work is long range understanding
of the narrative structure of movies. Instead of considering the entire movie, we propose to learn from the `key scenes' of the movie, providing a \texttt{condensed} look at the full storyline. To this end, we make the following three contributions: (i) We create the \texttt{Condensed Movies Dataset (CMD)} consisting of the key scenes from over 3K movies: each key scene is accompanied by a high level semantic description of the scene, character face-tracks, and metadata about the movie. The dataset is scalable, obtained automatically from YouTube, and is freely available for anybody to download and use. It is also an order of magnitude larger than existing movie datasets in the number of movies; 
(ii) We provide a deep network baseline for text-to-video retrieval on our dataset, combining character, speech and visual cues into a single video embedding; and finally (iii) We demonstrate how the addition of context from other video clips improves retrieval performance.

\end{abstract}

\section{Introduction}

\begin{figure}
\includegraphics[width=1\textwidth]{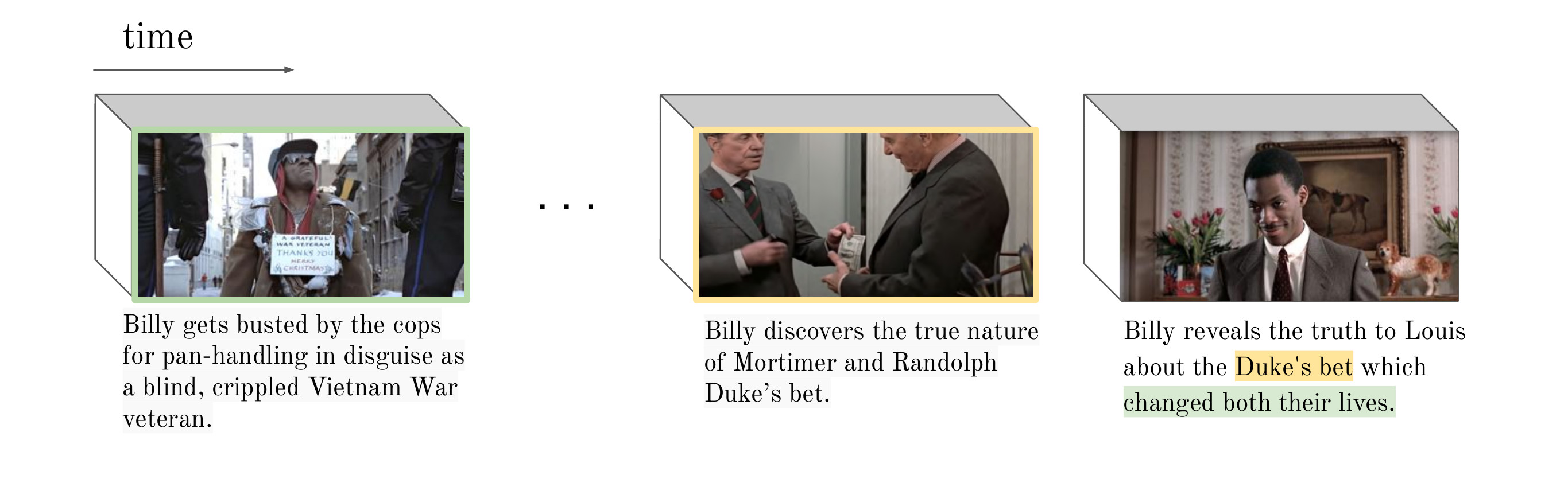}
\caption{\small{\textbf{Condensed Movies:} The dataset consists of the key scenes in a movie (ordered by time), together with high level semantic descriptions. Note how the caption of a scene (far right) is based on the knowledge of past scenes in the movie -- one where the Dukes exchange money to settle their bet (highlighted in yellow), and another scene showing their lives before the bet, homeless and pan-handling (highlighted in green).}}
\label{fig:teaser}
\vspace{-0.5em}
\end{figure}

Imagine you are watching the movie \textit{`Trading Places'}, and you want to instantly fast forward to a scene, one where `Billy reveals the truth to Louis about the Duke’s bet, a bet which changed both their lives'. In order to solve this task automatically, an intelligent system would need to watch the movie up to this point, have knowledge of Billy, Louis and the Duke's identities, understand that the Duke made a bet, and know the outcome of this bet (Fig.~\ref{fig:teaser}).
This high level understanding of the movie narrative requires knowledge of the characters' identities, their relationships, motivations and conversations, and ultimately their behaviour. Since movies and TV shows can provide an ideal source of data to test this level of story understanding, there have been a number of movie related datasets and tasks proposed by the computer vision community~\cite{tapaswi2014storygraphs,rohrbach2017movie,MSA,moviegraphs,huang2020movienet}. 

However, despite the recent proliferation of movie-related datasets, high level semantic understanding of human narratives still remains a challenging task. There are a number of reasons for this lack of progress: (i) semantic annotation is expensive and challenging to obtain, inherently restricting the size of current movie datasets to only hundreds of movies, and often, only part of the movie is annotated in detail~\cite{tapaswi2014storygraphs,rohrbach2017movie,MSA}; (ii) movies are very long (roughly 2 hours) and video architectures struggle to learn over such large timescales; (iii) there are legal and copyright issues surrounding a majority of these datasets~\cite{tapaswi2014storygraphs,MSA}, which hinder their widespread availability and adoption in the community; and finally (iv) the subjective nature of the task makes it difficult to define objectives and metrics~\cite{** need a ref here but the `right questions' paper has nothing to do with subjective **}.  

A number of different works have recently creatively identified that  certain domains of videos, such as
narrated instructional videos~\cite{miech2019howto100m,tang2019coin,zhou2018towards} and
lifestyle vlogs~\cite{ignat2019identifying,fouhey2018lifestyle} are available in large numbers on YouTube and are a good source of supervision for video-text models as the speech describes the video content. In a similar spirit, videos from the MovieClips channel on YouTube\footnote{https://www.youtube.com/user/movieclips}, which contains the key scenes or clips from numerous movies, are also accompanied by a semantic text description describing the content of each clip. 

Our first objective in this paper is to curate a dataset, suitable for learning and evaluating long range narrative structure understanding, from the available video clips and associated annotations of the MovieClips channel. To this end, we curate a dataset of `condensed' movies, called the Condensed Movie Dataset (CMD) which provides a \textit{condensed} snapshot into the entire storyline of a movie. In addition to just the video, we also download and clean the high level semantic descriptions accompanying each key scene that describes characters, their motivations, actions, scenes, objects, interactions and relationships. We also provide labelled face-tracks of the principal actors (generated automatically), as well as the metadata associated with the movie (such as cast lists, synopsis, year, genre). Essentially, all the information required to (sparsely) generate a MovieGraph~\cite{moviegraphs}. The dataset consists of over 3000 movies. 

Previous work on video retrieval and video understanding has largely
treated video clips as independent entities, divorced from their context~\cite{rohrbach2017movie,xu2016msr,anne2017localizing}.
But this is not how movies are understood: the meaning and significance of a scene depends on its
relationship to previous scenes.
This is true also of TV series, where one episode 
depends on those leading up to it (the season arc); and even an online
tutorial/lesson can refer to previous tutorials. These contextual videos are beneficial and sometimes even
necessary for complete video understanding. 

Our second objective is to explore the role of context  in enabling video retrieval.
We define a text-to-video retrieval task on the CMD, and extend
the
popular Mixture of Embedding Experts model~\cite{miech18learning}, that
can learn from the subtitles, faces, objects, actions and scenes, by
adding a \textit{Contextual Boost Module} that introduces information from past
and future clips. Unlike other movie related tasks -- e.g.\ text-to-video retrieval
on the LSMDC dataset~\cite{rohrbach2017movie} or graph retrieval on
the MovieQA~\cite{MovieQA} dataset that ignore identities, we also
introduce a character embedding module which allows the model to reason about the identities of characters present in each clip and description.
Applications of this kind of story-based retrieval include semantic search and indexing of movies as well as intelligent fast forwards. The CMD dataset can also be used for semantic video summarisation and automatic description of videos for the visually impaired (Descriptive Video Services (DVS) are currently available at a huge manual cost).

Finally, we also show preliminary results for aligning the semantic captions to the plot summaries of each movie, which places each video clip in the larger context of the movie as a whole. Data, code, models and features can be found at \url{https://www.robots.ox.ac.uk/~vgg/research/condensed-movies/}.

\section{Related Work}
\label{sec:related}

\noindent\textbf{Video Understanding from Movies:}
There is an increasing effort to develop video understanding techniques that go beyond action classification from cropped, short temporal snippets~\cite{kay2017kinetics,gu2018ava,monfort2019moments}, to learning from longer, more complicated videos that promise a higher level of abstraction~\cite{sener2015unsupervised,alayrac2016unsupervised,miech2019howto100m,sun2019videobert}. Movies and TV shows provide an ideal test bed for learning long-term stories, leading to a number of recent datasets focusing exclusively on this domain~\cite{tapaswi2014storygraphs,MovieQA,rohrbach2017movie,MSA}. Early works, however, focused on using film and TV to learn human identity~\cite{Everingham06a,naim2016aligning,cour2009learning,Sivic09,tapaswi2012knock,huang2020caption} or human actions~\cite{bojanowski2013finding,duchenne2009automatic,laptev2008learning,marszalek2009actions,Nagrani_2020_CVPR} from the scripts or captions accompanying movies. Valuable recent works have proposed story-based tasks such as the visualisation and grouping of scenes which belong to the same story threads~\cite{ercolessi2012stoviz,rao2020local}, the visualisation of TV episodes as a chart of character interactions~\cite{tapaswi2014storygraphs}, and more recently, the creation of more complicated movie graphs (MovieGraphs~\cite{moviegraphs} is the most exhaustively annotated movie dataset to date). Such graphs have enabled explicit learning of interactions and relationships~\cite{kukleva2020interactions} between characters. This requires understanding multiple factors such as human communication, emotions, motivation, scenes and other factors that affect behavior. There has also been a recent interest in evaluating story understanding through visual question answering~\cite{MovieQA} and movie scene segmentation~\cite{rao2020local}. In contrast, we propose to evaluate story understanding through the task of text-to-video retrieval, from a set of key scenes in a movie that condense most of the salient parts of the storyline. Unlike retrieval through a complex graph~\cite{moviegraphs}, retrieval via text queries can be a more intuitive way for a human to interact with an intelligent system, and might help avoid some of the biases present inherently in VQA datasets~\cite{bjasani2019movieqa}.\\
\noindent\textbf{Comparison to other Movie Datasets:}
Existing movie datasets often consist of short clips spanning entire,
full length movies (which are subject to copyright and difficult for
public release to the community). All such datasets also depend on
exhaustive annotation, which limit their scale to hundreds of
movies. Our dataset, in contrast, consists of only the key scenes from
movies matched with high quality, high level semantic descriptions,
allowing for a condensed look at the entire storyline. A comparison of
our dataset to other datasets can be seen in Table~\ref{tab:compare}.
\begin{table}
\vspace{-2em}
\setlength{\tabcolsep}{6pt}
\caption{\small{Comparison to other movie and TV show datasets. For completeness, we also compare to datasets that \textit{only}  have character ID or action annotation. `Free' is defined here as accessible online at no cost at the time of writing. *Refers to number of TV shows. }}
\centering
\footnotesize
\begin{tabular}{lrrcc}

\toprule
            & \multicolumn{1}{c}{\#Movies} & \multicolumn{1}{c}{\#Hours} &
            Free
            & Annotation Type \\ 
\midrule
Sherlock \cite{Nagrani17b} & 1* & 4 &   & Character IDs \\
TVQA\cite{lei2019tvqa} & 6* & 460 & & VQA \\
AVA \cite{gu2018ava} & 430 & 107.5 & \checkmark  & Actions only\\
MovieGraphs \cite{moviegraphs} & 51 & 93.9 &  & Descriptions, graphs\\
MovieQA (video)\cite{MovieQA}   & 140   & 381  &   & VQA \\
MovieScenes \cite{rao2020local} & 150 & 250 & & Scene segmentations \\
LSMDC \cite{rohrbach2017movie} & 202 & 158 & & Captions\\
MSA \cite{MSA} & 327 & 516 &   & Plots\\
MovieNet\cite{huang2020movienet} & 1,100 & 2,000 & & Plots, action tags, character IDs \\
\midrule
CMD (Ours)  & \textbf{3,605} & \textbf{1,270} &  \checkmark & \begin{tabular}[c]{@{}c@{}}Descriptions, metadata,\\  character IDs, plots
\end{tabular} \\
\bottomrule

\vspace{-0.5em}
\end{tabular}\label{tab:compare}
\vspace{-1em}

\end{table}

\noindent\textbf{Text-to-Video Retrieval:}

A common approach for learning visual embeddings from natural language supervision is to learn a joint embedding space where visual and textual cues are adjacent if they are semantically similar~\cite{miech18learning,Liu19a}. Most of these works rely on manually annotated datasets in which descriptive captions are collected for short, isolated video clips, with descriptions usually focusing on low-level visual content provided by annotators~\cite{rohrbach2017movie,anne2017localizing,xu2016msr}. 
For example LSMDC~\cite{rohrbach2017movie}, which is created from
DVS, contains mostly low-level
descriptions of the visual content in the scene, e.g.\ `Abby gets in
the basket', unlike the descriptions in our dataset.

Most similar to our work is~\cite{tapaswi2015aligning}, which obtains story level descriptions for shots in full movies, by aligning plot sentences to shots, and then attempting video retrieval. This, however, is challenging because often there is no shot that matches a plot sentence perfectly, and shots cover very small timescales. Unlike this work our semantic descriptions are more true to the clips themselves. \\

\noindent\textbf{Temporal Context:}
The idea of exploiting surrounding context has been explored
by~\cite{krishna2017dense}, for the task of video
captioning, and by~\cite{Wu_2019_CVPR} for video understanding.
Krishna {\it et al.}~\cite{krishna2017dense} introduces a new captioning module that uses
contextual information from past and future events to jointly describe
all events, however this work focuses on short term context (few seconds
before and after a particular clip). 
Wu {\it et al.}~\cite{Wu_2019_CVPR} go further, and introduce a feature bank architecture that can use
contextual information over several minutes, demonstrating  the performance improvements that results.
Our dataset provides the opportunity to extend such feature banks (sparsely) over an entire movie.

\section{Condensed Movie Dataset}
\label{sec:dataset}
\begin{figure}
     \centering
     \begin{subfigure}[b]{0.99\textwidth}
         \centering
        \includegraphics[width=1\textwidth]{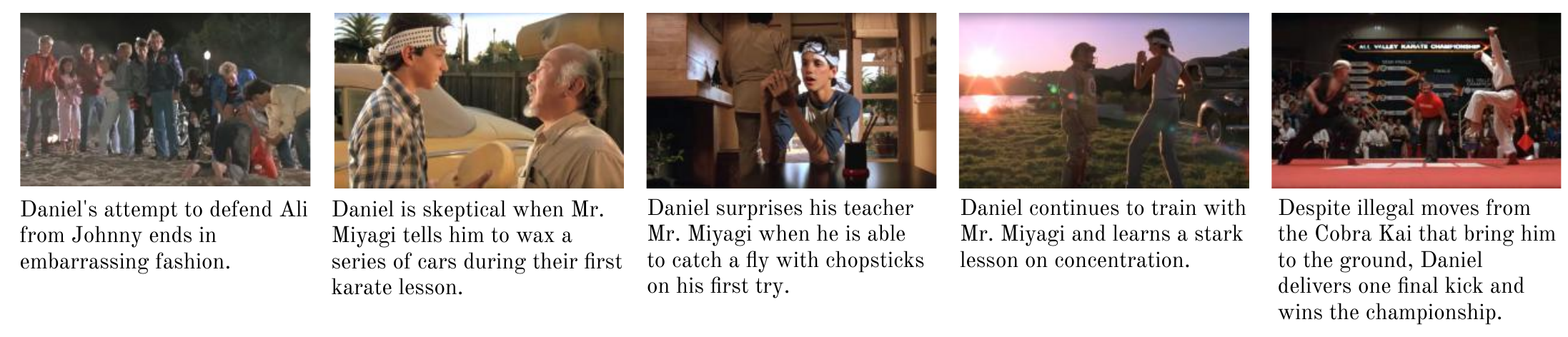}
     \end{subfigure}
    \begin{subfigure}[b]{1\textwidth}
        \includegraphics[width=1\textwidth]{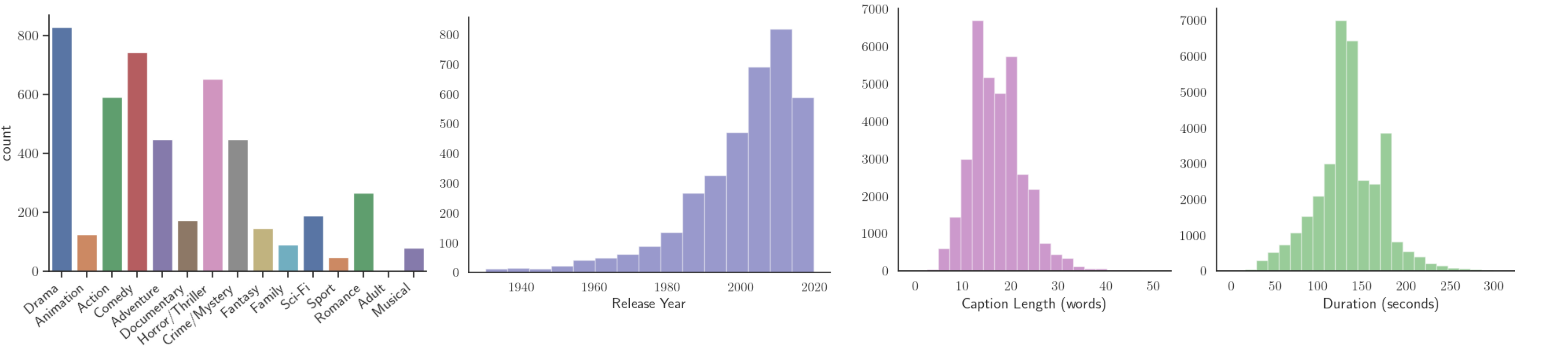}
    \end{subfigure}
    \begin{subfigure}[b]{0.88\textwidth}
         \centering
        \includegraphics[width=1\textwidth]{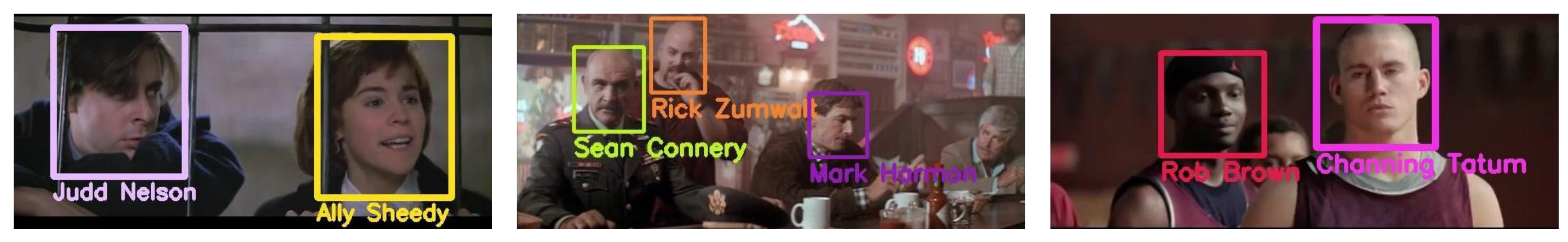}
     \end{subfigure}
\caption{\small{\textbf{The Condensed Movie Dataset (CMD)}. \textit{Top:} Samples of clips and their corresponding captions from \textit{The Karate Kid (1984)} film. In movies, as in real life, situations follow from other situations and the combination of video and text tell a concise story. Note: Every time a character is mentioned in the description, the name of the actor is present in brackets. We remove these from the figure in the interest of space. \textit{Middle, from left to right:} Histogram of movie genres, movie release years, description length and duration of video clips. Best viewed online and zoomed in. \textit{Bottom:} Example face-tracks labelled with the actor's name in the clips. These labels are obtained from cast lists and assigned to face-tracks using our automatic labelling pipeline.}}
\label{fig:three_graphs}
\vspace{-0.5em}
\end{figure}
\vspace{-0.5em}

\begin{figure}
     \centering
        \includegraphics[width=1\textwidth]{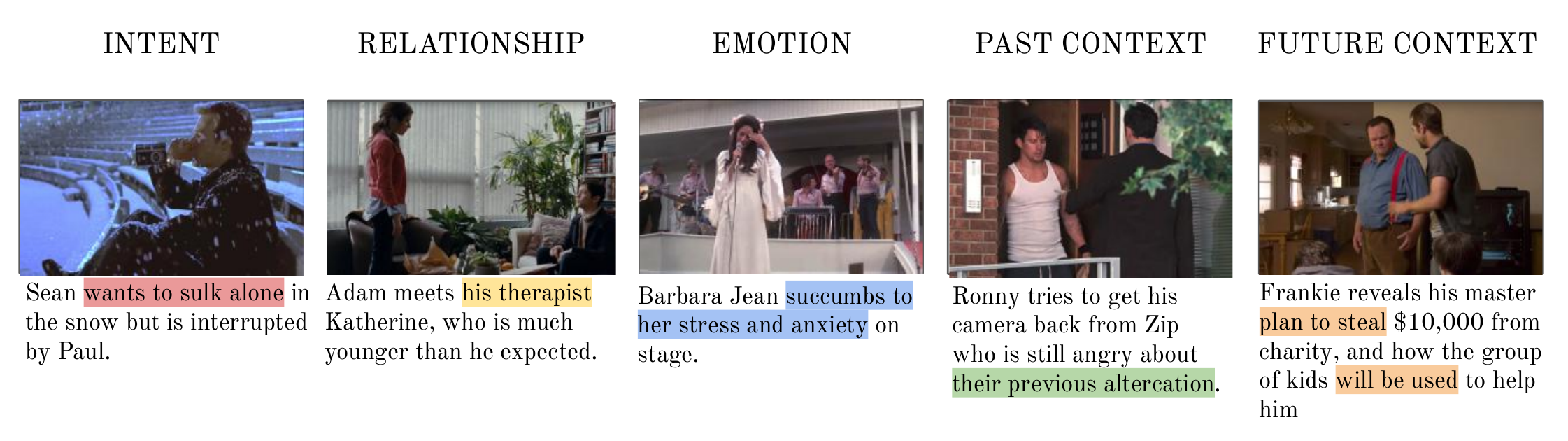}
\caption{\small{\textbf{Semantic descriptions:} Examples of high level semantic descriptions accompanying each video clip in our dataset (note: actor names are removed to preserve space). Our semantic descriptions cover a number of high level concepts, including intent/motivation, relationships, emotions and attributes, and context from surrounding clips in the storyline.}}
\label{fig:descriptions}
\vspace{-0.5em}
\end{figure}

We construct a dataset to facilitate machine understanding of narratives in long movies.  
Our dataset has the following key properties: \\
\noindent\textbf{(1) Condensed Storylines:} 
The video data consists of over 33,000 clips from 3,600 movies (see Table \ref{tab:cap_compare}). For each movie there is a set of ordered clips (typically 10 or so) covering the salient parts of the film (examples can be seen in Fig. \ref{fig:three_graphs}, top row). Each around two minutes in length, the clips contain the same rich and complex story as full-length films but an order of magnitude shorter.
The distribution of video lengths in our dataset can be seen in Fig. \ref{fig:three_graphs} -- with just the key scenes, each movie has been condensed into roughly 20 minutes each. Each clip is also accompanied by a high level description focusing on intent, emotion, relationships between characters and high level semantics (Fig. ~\ref{fig:three_graphs} and \ref{fig:descriptions}). Compared to other video-text datasets, our descriptions are longer, and have a higher lexical diversity~\cite{McCarthy2010} (Table \ref{tab:cap_compare}). We also provide face-tracks and identity labels for the main characters in each clip (Fig. ~\ref{fig:three_graphs}, bottom row). \\
\noindent\textbf{(2) Online Longevity and Scalability:} All the videos are obtained from the licensed, freely available YouTube channel: MovieClips by Fandango. We note that a common problem plaguing YouTube datasets today~\cite{caba2015activitynet,gu2018ava,kay2017kinetics,Nagrani19} is the fast shrinkage of datasets as user uploaded videos are taken down by users (over 15\% of Kinetics-400~\cite{kay2017kinetics} is no longer available on YouTube at the time of writing, including videos from the eval sets). We believe our dataset has longevity due to the fact that the movie clips on the licensed channel are rarely taken down from YouTube. Also, this is an actively growing YouTube channel as new movies are released and added. Hence there is a potential to continually increase the size of the dataset. We note that from the period of 1st Jan 2020, to 1st September 2020, only 0.3\% of videos have been removed from the YouTube channel, while an additional 2,000 videos have been uploaded, resulting in a dataset growth of 5.8\% over the course of 9 months.  \\

\begin{table}[]
\vspace{-2em}
\setlength{\tabcolsep}{6pt}
\caption{\small{Comparison to other video text retrieval datasets. MTLD is the Measure of Textual Lexical Diversity~\cite{mtld} for all of the descriptions in the dataset.}}
\centering

\begin{tabular}{lrccc}

\toprule
Dataset & \begin{tabular}[c]{@{}c@{}}\#Videos/\#Clips\end{tabular} & \begin{tabular}[c]{@{}c@{}}Median caption\\ len. (words)\end{tabular}& MTLD & \begin{tabular}[c]{@{}c@{}}Median clip\\ len. (secs)\end{tabular} \\ 
\midrule
MSVRTT \cite{xu2016msr} & 7,180/10,000 & 7 & 26.9 & 15 \\
DiDemo \cite{anne2017localizing} & 10,464/26,892 & 7 & 39.9 & 28\\
LSMDC \cite{rohrbach2017movie} & 200/118,114 & 8 & 61.6 & 5\\
CMD (Ours)  & 3,605/33,976 & \textbf{18} & \textbf{89.1} & \textbf{132} \\
\bottomrule

\end{tabular}\label{tab:cap_compare}

\end{table}

\subsection{Dataset Collection Pipeline}
\label{sec:dataset_collection}
In this section we describe the dataset collection pipeline. \\
\noindent\textbf{Videos and  Descriptions:}
Raw videos are downloaded from YouTube. Each video is accompanied by an outro at the end of the clip which contains some advertising and links to other movies. This is automatically removed by using the observation that each outro has a consistent length of either 10s (if the clip is uploaded before May 2017) or 30s if uploaded after. Approximately 1,000 videos from the channel were manually excluded from the dataset because they contained low quality descriptions or did not contain scenes from a movie. For each video, we also download the YouTube closed captions, these are a mix of high quality, human generated subtitles and automatic captions. Closed captions are missing for 36.7\% of the videos.
The MovieClips channel also provides a rich and high level description with each video, which we extract, clean (removing the movie title, links and advertising) and verify manually.
We note that the videos also contain a watermark, usually at the bottom left of the frame. These can be easily cropped from the videos. \\
\noindent\textbf{Metadata:}
For each clip, we identify its source movie by parsing the movie title from the video description and, if available, the release year (since many movies have non-unique titles). The title and release year are queried in the IMDb search engine to obtain the movie's IMDb ID, cast list and genre. IMDb identification enables correspondence to other popular movie datasets~\cite{MovieQA,lsmdc}. Plot synopses were gathered by querying the movie title and release year in the Wikipedia search engine and extracting text within the `Plot' section of the top ranked entry. For each movie we include: (i) the movie description (short, 3-5 sentences), accompanying the video clips on the MovieClips YouTube channel; (ii) Wikipedia plot summaries (medium, 30 sentences); and (iii) IMDB plot synopses (long, 50+ sentences). 
\\
\noindent\textbf{Face-tracks and Character IDs:}
We note that often character identities are the focal point of any storyline, and many of the descriptions reference key characters. In a similar manner to~\cite{Nagrani17b}, we use face images downloaded from search engines to label detected and tracked faces in our dataset. Our technique involves the creation of a character embedding bank (CEB) which contains a list of characters  (obtained from cast lists), and a corresponding embedding vector obtained by passing search engine image results through a deep CNN model pretrained on human faces \cite{Cao18}. Character IDs are then assigned to face-tracks in the video dataset when the similarity between the embeddings from the face tracks and the embeddings in the CEB (using cosine similarity) is above a certain threshold.  This pipeline is described in detail in the appendix, Sec.~\ref{sec:app:cidpipeline}. We note that this is an automatic method and so does not yield perfect results, but a random manual inspection shows that it is accurate 96\% of the time. Ultimately, we are able to recognize 8,375 different characters in 25,760 of the video clips.\\

\subsection{Story Coverage}

To quantitatively measure the amount of the story covered by movie clips in our dataset, we randomly sample 100 movies and manually aligned the movie clips (using the descriptions as well as the videos) to Wikipedia plot summaries (the median length of which is 32 sentences). We found that while the clips totalled only \textbf{15\%} of the full-length movie in time duration, they cover \textbf{44\%} of the full plot sentences, suggesting that the clips can indeed be described as key scenes. In addition, we find that the movie clips span a median range of \textbf{85.2\%} of the plot, with the mean midpoint of the span being \textbf{53\%}. 
We further show the distribution of clip sampling in Fig. \ref{fig:plot_coverage} in the appendix, Sec.~\ref{sec:app:plotalign} and find that in general there is an almost uniform coverage of the movie. While we focus on a baseline task of video-text retrieval, we also believe that the longitudidal nature of our dataset will encourage other tasks in long range movie understanding.

\section{Text-to-Video Retrieval} \label{story_based_retrieval}
In this section we provide a baseline task for our dataset -- the task of text-to-video retrieval. The goal here is to retrieve the correct `key scene' over all movies in the dataset, given just the high level description. Henceforth, we use the term `video clip' to refer to one key scene, and `description' to refer to the high level semantic text accompanying each video clip. In order to achieve this task, we learn a common embedding space for each video and the description accompanying it.
More formally, if $V$ is the video and $T$ is the description, we learn embedding functions $f$ and $g$ such that the similarity $s=\langle f(V), g(T)\rangle$ is high only if $T$ is the correct semantic description for the video $V$. Inspired by previous works that achieve state-of-the-art results on video retrieval tasks~\cite{miech18learning,Liu19a}, we encode each video as a combination of different streams of descriptors. Each descriptor is a semantic representation of the video learnt by individual experts (that encode concepts such as scenes, faces, actions, objects and the content of conversational speech from subtitles). 

\begin{figure}
\vspace{-0.5em}
     \centering
        \includegraphics[width=1\textwidth]{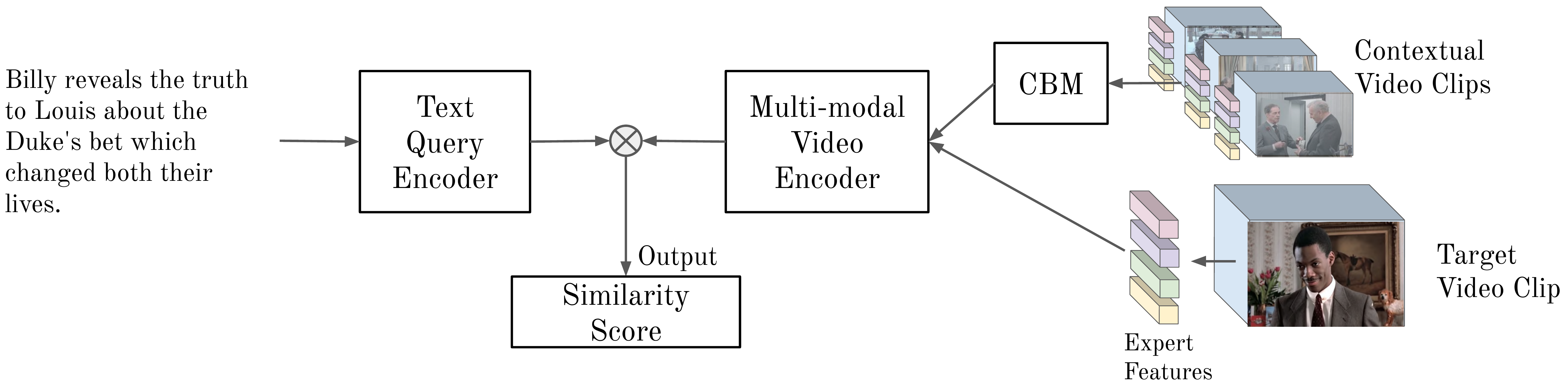}
\caption{\small{\textbf{Model architecture:} An overview of text-to-video retrieval with our Contextual Boost module (CBM) that computes a similarity score between a query sentence $T$ and a target video. CBM receives contextual video features (which are previous clips from the same movie) to improve the multimodal encoding of the target video clip. The expert features are extracted using pre-trained models for speech, motion, faces, scenes and objects.}}
\label{fig:model}
\vspace{-2em}
\end{figure}

Inspired by~\cite{miech18learning}, we base our network architecture on a mixture of `expert' embeddings model, wherein a separate model is learnt for each expert, which are then combined in an end-to-end trainable fashion using weights that depend on the input caption. This allows the model to learn to increase
the relative weight of motion descriptors for input captions concerning human actions,
or increase the relative weight of face descriptors for input captions that require detailed
face understanding. 
We also note, however, that often the text query not only provides clues as to which expert is more valuable, but also whether it is useful to pay attention to a previous clip in the movie, by referring to something that happened previously, eg. `Zip is \textit{still} angry about their \textit{previous altercation}'. Hence we introduce a Contextual Boost module (CBM), which allows the model to learn to increase the relative weight of a past video feature as well. A visual overview of the retrieval system with the CBM can be seen in Fig. \ref{fig:model}. In regular movie datasets, the space of possible previous clips can be prohibitively large~\cite{tapaswi2015aligning}, however this becomes feasible with our \textit{Condensed Movies} dataset.

Besides doing just \textit{cross-movie} retrieval, we also adapt our model to perform \textit{within-movie} retrieval. We note that characters are integral to a storyline, and hence for the case of within-movie retrieval, we introduce a character module, which computes a weighted one-hot vector for the characters present in the description query and another for each video clip in the dataset. We note that for cross-movie retrieval, the retrieval task becomes trivial given the knowledge of the characters in each movie, and hence to make the task more challenging (and force the network to focus on other aspects of the story), we remove the character module for this case.

\vspace{-1em}

\subsection{Model Architecture}
\subsubsection{Expert Features.}
Stories in movies are communicated through many modalities including (but not limited to) speech, body language, facial expressions and actions. Hence we represent each input video $V$ with $K$ different expert streams (in our case, $K=5$ --  face, subtitles, objects, motion and scene, but our framework can be extended to more experts as required).

 Each input stream is denoted as $I_i$, where $i=1,...,K$. Adopting the approach proposed by~\cite{miech18learning}, we first aggregate the descriptors of each input stream over time, using a temporal aggregation module (see Sec.~\ref{sec:experiments} for details), and the resulting time-aggregated descriptor is embedded using a gated embedding module (for the precise details of the gated embedding module, please see~\cite{miech18learning}). We then finally project each embedding to a common dimension $D$ using a fully connected layer, giving us one expert embedding $E_{V_i}$ for each input stream $i$.
Hence the final output is of dimensions $K \times D$.

\subsubsection{Text Query Encoder.}
The query description input is a sequence of BERT word embeddings~\cite{devlin2018bert} for each input sentence. These individual word embedding vectors are then aggregated into a single vector $h(T)$ representing 
the entire sentence using a NetVLAD~\cite{arandjelovic2016netvlad} aggregation module. This vector $h(T)$, is used to predict the mixture weights (described in the next section). We project $h(T)$ to the same dimensions as the video expert features using the same gated embedding module followed by a fully connected layer as for the video experts (described above), once for each input source $i$, giving us expert embeddings $E_{T_i}$. Hence the final output is also of dimensions $K \times D$.

\subsubsection{Contextual Boost Module.}
In both ~\cite{miech18learning} and ~\cite{Liu19a}, the resulting expert embeddings $E_{V_i}$ are then weighted using normalised weights $w_i(T)$ estimated from the text description $T$. The final similarity score $s$ is obtained by a weighted combination of the similarity scores $s_i(E_{T_i},E_{V_i})$ between the embeddings $E_{T_i}$ of the query sentence $T$ and the expert embeddings $E_{V_i}$ (obtained from the input video descriptors $I_i$). More formally, this is calculated as: 
\begin{align}
    s(T,V) = \sum_{i=1}^{K} w_i(T)s_i(E_{T_i},E_{V_i}), & & \textnormal{where} & & w_i(T) =  \frac{e^{h(T)^{\intercal}a_i}}{\sum_{j=1}^{K}e^{h(T)^{\intercal}a_j}}
    \label{eqn:weights}
\end{align}
where $s_i$ is the scalar product, $h(T)$ is the aggregated text query representation described above and $a_i$, $i = 1,..,.K$ are learnt parameters used to obtained the mixture weights. 

In this work, however, we extend this formulation in order to incorporate past context into the retrieval model. We would like the model to be able to predict weights for combining experts from previous clips -- note we treat each expert separately in this formulation. For example, the model might want to heavily weight the subtitles from a past clip, but down-weight the scene representation which is not informative for a particular query. More formally, given the total number of clips we are encoding to be $N$, we modify the equation above as: 
\begin{align}
    s(T,V) = \sum_{n=1}^{N}\sum_{i=1}^{K} w_{i,n}(T)s_{i,n}(E_{T_i},E_{V_{i,n}}), \\
    w_{i,n}(T) =  \frac{e^{h(T)^{\intercal}a_{i,n}}}{\sum_{m=1}^{N}\sum_{j=1}^{K}e^{h(T)^{\intercal}a_{j,m}}}.
    \label{eqn:weights1}
\end{align}
Hence instead of learning K scalar weights $a_i$, $i = 1,...,K$ as done in~\cite{miech18learning} and~\cite{Liu19a}, we learn $K\times N$ scalar weights $a_{i,n}$, $i = 1,...,K$, $n = 1,...,N$ to allow combination of experts from additional clips. 

\subsubsection{Dealing with missing streams.}
We note that these experts might be missing for certain videos, e.g.\ subtitles are not available for all videos and some videos do not have any detected faces. When expert features are missing, we zero-pad the missing experts and compute the similarity score. This is the standard procedure followed by existing retrieval methods using Mixture of Embedding Experts models~\cite{miech18learning,Liu19a}. The similarity score is calculated only from the available experts by re-normalising the mixture weights to sum to one, allowing backpropagation of gradients only to the expert branches that had an input feature. We apply this same principle when dealing with missing video clips in the past, for example if we are training our model with $N=1$ past clips, for a video clip which is right at the start of the movie (has no past), we treat all the experts from the previous clip as missing so that the weights are normalised to focus only on the current clip.  

\subsubsection{Character Module.}
The character module computes the similarity between a vector representation of the character IDs mentioned in the query $y$ and a vector representation of the face identities recognised in the clip $x$. The vector representations are computed as follows:
For the query, we search for actor names in the text from the cast list (supplied by the dataset) and create a one-hot vector $y$ the same length as the cast list, where $y_i = 1$ if actor $i$ is identified in any face track in the video and $y_i = 0$ otherwise. For the face identities acquired in the face recognition pipeline (described earlier), we compare the following three methods: first, we encode a one-hot vector $x$ in a manner similar to the query character encoding.

While this can match the presence and absence of characters, it doesn't allow any weighting of characters based on their importance in a clip.

Hence inspired by~\cite{Tapaswi2015_Book2Movie}, we also propose a second method (``track-frequency normalised''), where $x_{i}$ is the number of face tracks for identity $i$. Lastly, in ``track length normalised'', our vector encodes the total amount of time a character appears in a clip i.e. $x_i$ is the sum of all track lengths for actor $i$, divided by the total sum of all track lengths in the clip. The performances of the three approaches are displayed and discussed in Table~\ref{tab:intra-results} and Section~\ref{sec:experiments} respectively.
The character similarity score $s_C = \langle y,x\rangle$ is then modulated by its own scalar mixture weight $w_C(T)$ predicted from $h(T)$ (as is done for the other experts in the model). This similarity score is then added to the similarity score obtained from the other experts to obtain the final similarity score, i.e.  $s(T,V) = \sum_{i=1}^{K} w_i(T)s_i(E_{T_i},E_{V_i}) + w_C(T)s_C(T,V)$. \\
\noindent\textbf{Training Loss.}
As is commonly done for video-text retrieval tasks, we minimise the Bidirectional Max-margin Ranking Loss~\cite{socher2014grounded}.

\section{Experiments}
\label{sec:experiments}
\subsection{Experimental Set-up}
We train our model for the task of cross-movie and within-movie retrieval. The dataset is split into disjoint training, validation and test sets by movie, so that there are no overlapping movies between the sets. The dataset splits can be seen in Table~\ref{tab:splits}. We report our results on the \textit{test set}
using standard retrieval
metrics including median rank (lower is better), mean
rank (lower is better) and R@K (recall at rank K—higher is better). \\
\noindent\textbf{Cross-movie Retrieval:}
 For the case of cross-movie retrieval, the metrics are reported over the entire test set of videos, i.e. given a text query, there is a `gallery' set of 6,581 possible matching videos
 (Table~\ref{tab:splits}). We report R@1, R@5, R@10, mean and median rank.\\
\noindent\textbf{Within-movie Retrieval:}
In order to evaluate the task of within-movie retrieval, we remove all movies that contain less than $5$ video clips from the dataset. For each query text, the possible gallery set consists only of the videos in the same  movie as the query. In this setting the retrieval metrics are calculated separately for each movie and then averaged over all movies. We report R@1, mean and median rank. 
\subsection{Baselines}
\noindent The \textbf{E2EWS} (End-to-end  Weakly  Supervised) is a cross-modal retrieval model trained by \cite{Miech_2020_CVPR} using weak supervision from a large-scale corpus of (100 million)  instructional videos (using speech content as the supervisory signal). We use the video and text encoders without any form of fine-tuning on Condensed Movies, to demonstrate the widely different domain of our dataset.\\
The \textbf{MoEE} (Mixture  of  Embedded  Experts)  model  proposed  by \cite{miech18learning} comprises a multi-modal video model in combination with a system of context gates that learn to fuse together different pretrained experts.  
\\
The \noindent\textbf{CE} model \cite{Liu19a} similarly learns a cross-modal embedding by fusing together a collection of pretrained experts to form a video encoder, albeit with pairwise relation network sub-architectures. It represents the state-of-the-art on several retrieval benchmarks. \\
\noindent\textbf{Context Boosting Module:} Finally, we report results with the addition of our Context Boosting module to both MoEE and CE. We use the fact that the video clips in our dataset are ordered by the time they appear in the movie, and encode previous and future `key scenes' in the movie along with every video clip using the CBM. An ablation on the number of clips encoded for context can be found in the supplementary material.  \\
\begin{table}[t]
 \vspace{-2.5mm}
\centering
\tabcolsep=0.19cm
\caption{\small{Training splits for cross-movie retrieval (left) and within-movie retrieval (right). For within-movie retrieval, we restrict the dataset to movies which have at least $5$ video clips in total.}}
\small
\begin{tabular}{lrrrr|rrrr}
\toprule
& \multicolumn{4}{c}{Cross-Movie} & \multicolumn{4}{c}{Within-Movie} \\
                      & \textsc{Train}   & \textsc{Val}    & \textsc{Test}   & \textsc{Total}                      & \textsc{Train}   & \textsc{Val}    & \textsc{Test}   & \textsc{Total}              \\
\midrule
 \#Movies              & 2,551      & 358     & 696     & 3,605  & 2,469      & 341     & 671     & 3,481             \\
 \#Video clips         & 24,047    & 3,348   & 6,581   & 33,976   & 23,963    & 3,315   & 6,581   & 33,859            \\
\bottomrule
\end{tabular}
\label{tab:splits}
\vspace{-1em}
\end{table}
We finally show the results of an ablation study demonstrating the importance of different experts for this task on the task of cross-movie retrieval. 

In the next sections, we first describe the implementation details of our models and then discuss quantitative and qualitative results. 

\subsection{Implementation Details}
\label{sec:implementation}
\noindent\textbf{Expert Features:}
In order to capture the rich content of a video, we draw on existing powerful representations for a number of different semantic tasks. These are first extracted at a frame-level, then aggregated by taking the mean to produce a single feature vector per modality per video. \\
\noindent\textbf{RGB object} frame-level embeddings of the visual data are generated with an SENet-154 model~\cite{hu2019squeeze} pretrained on ImageNet for the task of image classification. Frames are extracted at 25 fps, where each frame is resized to $224 \times 224$ pixels. Features collected have a dimensionality of $2048$. \\
\noindent\textbf{Motion} embeddings are generated using the I3D inception model~\cite{carreira2017quo} trained on Kinetics~\cite{kay2017kinetics}, following the procedure described by~\cite{carreira2017quo}.  \\
\noindent\textbf{Face} embeddings for each face track are extracted in three stages: (1) Each frame is passed through a dual shot face detector~\cite{li2019dsfd} (trained on the Wider Face dataset~\cite{yang2016wider}) to extract bounding boxes. (2) Each box is then passed through an SENet50~\cite{Hu18} trained on the VGGFace2 dataset~\cite{Cao18} for the task of face verification, to extract a facial feature embedding, which is L2 normalised. (3) A simple tracker is used to connect the bounding boxes temporally within shots into face tracks. Finally the embeddings for each bounding box within a track are average pooled into a single embedding per face track, which is again L2 normalised. The tracker uses a weighted combination of intersection over union and feature similarity (cosine similarity) to link bounding boxes in consecutive frames. \\
\noindent\textbf{Subtitles} are encoded using BERT embeddings~\cite{devlin2018bert} averaged across all words. \\
\noindent\textbf{Scene} features of 2208 dimensions are encoded using a DenseNet161 model~\cite{iandola2014densenet} pretrained on the Places365 dataset~\cite{zhou2017places}, applied to $224 \times 224$ pixel centre crops of frames extracted at 1fps.  \\
\noindent\textbf{Descriptions} are encoded using BERT embeddings, providing contextual word-level features of dimensions $W \times1024$ where W is the number of tokens. These are concatenated and fed to a NetVLAD layer to produce a feature vector of length of 1024 times the number of NetVLAD clusters for variable length word tokens. 

\noindent\textbf{Training details and hyperparameters:}
All baselines and CBM are implemented with PyTorch~\cite{paszke2017automatic}. Optimization is performed with Adam~\cite{Kingma2014AdamAM}, using a learning rate of 0.001, and a batch size of 32. The margin hyperparameter $m$ for the bidirectional ranking loss is set to a value of 0.121, the common projection dimension $D$ to 512, and the description NetVLAD clusters to $10$. For CBM, we select the number of past and future context videos to be N=3, ablations for hyperparameters and using different amounts of context are given in the supplementary material.
Training is stopped when the validation loss stops decreasing. 

\begin{table}
\vspace{-1.5em}
\caption{\small{Cross-movie text-video retrieval results on the CMD \textit{test} set of 6,581 video clips, with varying levels of context. Random weights refers to the MoEE model architecture with random initialization. We report Recall$@$k (higher is better), Median rank and Mean rank (lower is better).}}
\centering
\small
\setlength{\tabcolsep}{3pt}
\begin{tabular}{l|ccccc|}
\toprule
\multicolumn{1}{l|}{Method} & \multicolumn{1}{c}{Recall$@$1} & \multicolumn{1}{c}{Recall$@$5} & \multicolumn{1}{c}{Recall$@$10} & \multicolumn{1}{c}{Median Rank} & \multicolumn{1}{c}{Mean Rank} \\ 
\midrule
\multicolumn{1}{l|}{Random weights} & 0.0 & 0.1 & 0.2 & 3209 & \multicolumn{1}{r}{3243.5} \\
\multicolumn{1}{l|}{E2EWS \cite{Miech_2020_CVPR}} & 0.7 & 2.2 & 3.7 & 1130 & \multicolumn{1}{r}{1705.5}\\
\multicolumn{1}{l|}{CE \cite{Liu19a}} & 2.3 & 7.4 & 11.8 & 190 & \multicolumn{1}{r}{570.0} \\
\multicolumn{1}{l|}{MoEE \cite{miech18learning}} & 4.7 & 14.9 & 22.1 & 65 & \multicolumn{1}{r}{285.3} \\
\multicolumn{1}{l|}{CE + CBM (ours)} & 3.6 & 12.0 & 18.2 & 103 & \multicolumn{1}{r}{474.6} \\
\multicolumn{1}{l|}{\textbf{MoEE + CBM (ours)}}& \textbf{5.6} & \textbf{17.6} & \textbf{26.1} & \textbf{50} & \multicolumn{1}{r}{\textbf{243.9}}  \\
\bottomrule
\end{tabular}\label{tab:inter-results}
\vspace{-3em}
\end{table}

\begin{table}[]
\centering
\caption{\small{Within-Movie Retrieval results on the CMD test set. All movies with less than $5$ video clips are removed. Metrics are computed individually for each movie and then averaged (m-MdR and m-MnR refers to the mean of the median and mean rank obtained for each movie respectively). R@1 denotes recall@1. We show the results of $3$ different variations of embeddings obtained from the character module.}}
\small
\begin{tabular}{l|rrr}
\toprule
Method          & \multicolumn{1}{c}{m-R$@$1} & \multicolumn{1}{c}{m-MdR} & \multicolumn{1}{c}{m-MnR} \\ 
\midrule
Random weights  & 11.1                     & 5.32                    & 5.32  \\
MoEE            & 38.9                      & 2.20                    & 2.82  \\
MoEE + Character Module [one-hot] & 45.5                      & 1.91                    & 2.60   \\
MoEE + Character Module [track-len norm] & 46.2                      & 1.88                    & 2.53 \\               
MoEE + Character Module \textbf{[track-freq norm]} & \textbf{47.2}                      & \textbf{1.85}                    & \textbf{2.49}  \\
\bottomrule
\end{tabular}\label{tab:intra-results}
\vspace{-1.5em}
\end{table}
\vspace{-1em}

\begin{table}
\caption{\small{Expert ablations. The value of different experts in combination with a baseline for text-video retrieval (left) and (right) their cumulative effect (here Prev. denotes the experts used in the previous row). R@k: recall@k, MedR: median rank, MeanR: mean rank}}
\centering
\small
\setlength{\tabcolsep}{1pt}
\begin{tabular}{l|rrrrr|l|rrrrr}
\toprule
Experts & \multicolumn{1}{c}{R$@$1} & \multicolumn{1}{c}{R$@$5} & \multicolumn{1}{c}{R$@$10} & \multicolumn{1}{c}{MedR} & \multicolumn{1}{c|}{MeanR} & Experts & \multicolumn{1}{c}{R$@$1} & \multicolumn{1}{c}{R$@$5} & \multicolumn{1}{c}{R$@$10} & \multicolumn{1}{c}{MedR} & \multicolumn{1}{c}{MeanR} \\ \midrule
Scene & 0.8 & 3.2 & 5.9 & 329 & 776.3 & Scene & 0.8 & 3.2 & 5.9 & 329 & 776 \\
Scene+Face & 3.7 & 12.7 & 19.7 & 100 & 443.1 & Prev.+Face & 3.7 & 12.7 & 19.7 & 100 & 443.1 \\
Scene+Obj & 1.0 & 4.6 & 8.0 & 237 & 607.8 & Prev.+ Obj & 3.9  & 13.1 & 20.5 & 79  & 245.5  \\
Scene+Action & 1.9 & 6.4 & 10.5 & 193 & 575.0 & Prev.+ Action & 4.0 & 14.0 & 20.4 & 78 & 233.3 \\
Scene+Speech & 2.3 & 8.3 & 12.4 & 165 & 534.7 & Prev.+Speech & 5.6 & 17.6 & 26.1 & 50 & 243.9\\
\bottomrule
\end{tabular}\label{tab:ablation}
\vspace{-0.5em}

\end{table}
\vspace{-0.5em}


\begin{figure}
    \centering
    \includegraphics[width=\textwidth]{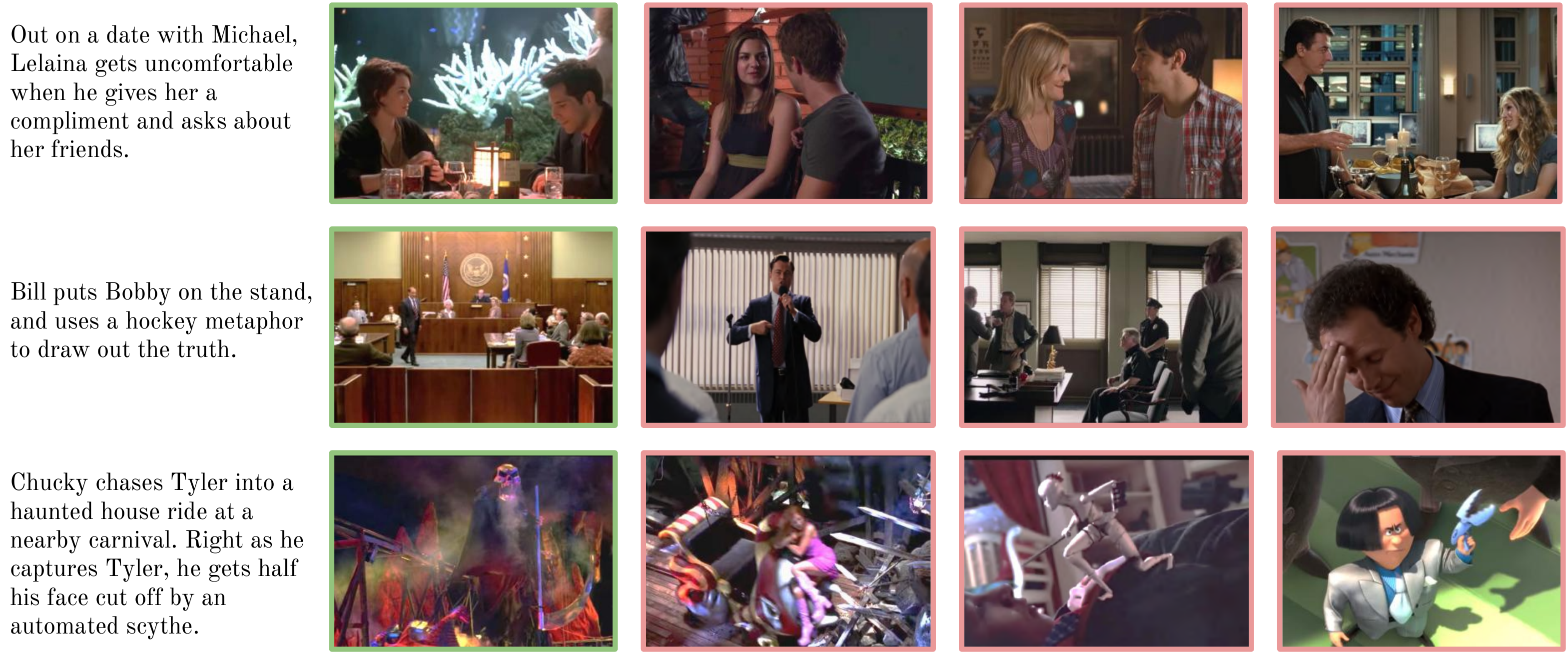}
    \caption{\small{\textbf{Qualitative results of the MoEE+CBM model for cross-movie retrieval.} On the left, we provide the input query, and on the right, we show the top 4 video clips retrieved by our model on the CMD \textit{test set}. A single frame for each video clip is shown. The matching clip is highlighted with a green border, while the rest are highlighted in red (best viewed in colour). Note how our model is able to retrieve semantic matches for situations (row 1: male/female on a date), high level abstract concepts (row 2: the words `stand' and `truth' are mentioned in the caption and the retrieved samples show a courtroom, men delivering speeches and a policeman's office) and also notions of violence and objects (row 3: scythe). }}
    \label{fig:qual}
\end{figure}
\subsection{Results}
\label{sec:results}

Results for cross-movie retrieval can be seen in Table~\ref{tab:inter-results}. E2EWS performs poorly, illustrating the domain gap between CMD and generic YouTube videos from HowTo100M.   
Both the CE and MoEE baselines perform much better than random, demonstrating that story-based retrieval is achievable on this dataset. We show that the Contextual Boost module can be effectively used in conjunction with existing video retrieval architectures, improving performance for both CE and MoEE, with the latter being the best performing model.
Results for within-movie retrieval can be seen in Table \ref{tab:intra-results}. We show that adding in the character module provides a significant boost (almost a 10\% increase in Recall@1 compared to the MoEE without the character module), with the best results obtained from normalizing the character embeddings by the track frequency. 
The value of different experts is assessed in Table~\ref{tab:ablation}. Since experts such as subtitles and face are missing for many video clips, we show the performance of individual experts combined with the `scene' expert features, the expert with the lowest performance that is consistently available for all clips (as done by~\cite{Liu19a}). In Table~\ref{tab:ablation}, right, we show the cumulative effect of adding in the different experts. The highest boosts are obtained from the face features and the speech features, as expected, since we hypothesize that these are crucial for following human-centric storylines. We show qualitative results for our best cross-movie retrieval model (MoEE + CBM) in Fig.~\ref{fig:qual}.

\vspace{-1em}
\section{Plot Alignment}
\begin{figure}
\vspace{-2em}
    \centering
    \includegraphics[width=\textwidth]{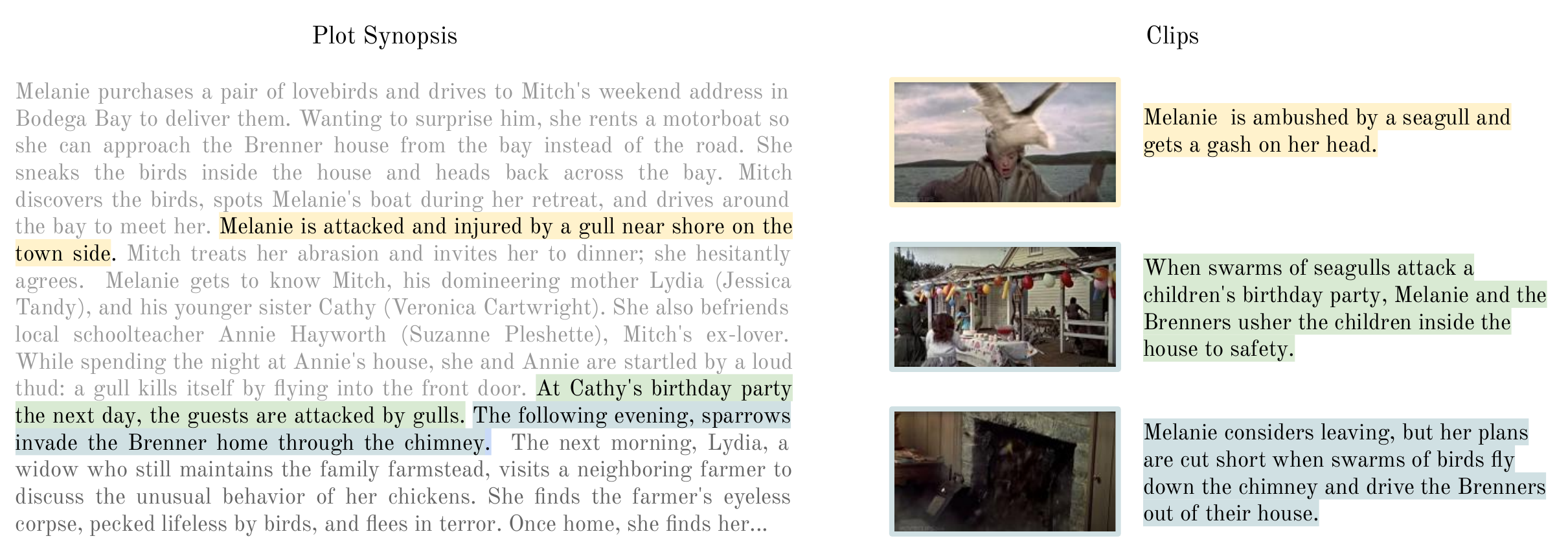}
    \caption{\small{A sample Wikipedia movie plot summary (left) aligned with an ordered sample of clips and their descriptions (right). The alignment was achieved using Jumping Dynamic Time Warping \cite{jdtw} of sentence-level BERT embeddings, note how the alignment is able to skip a number of peripheral plot sentences.}}
    \label{fig:plot_desc_alignment}
    \vspace{-0.5em}

\end{figure}

\vspace{-0.5em}
A unique aspect of the Condensed Movies Dataset is the story-level captions accompanying the ordered key scenes in the movie. Unlike existing datasets~\cite{rohrbach2017movie} that contain low level visual descriptions of the visual content, our semantic captions capture key plot elements.
To illustrate the new kinds of capabilities afforded by this aspect, we align the video descriptions to Wikipedia plot summary sentences using Jumping Dynamic Time Warping~\cite{jdtw} of BERT sentence embeddings. This alignment allows us to place each video clip in the global context of the larger plot of the movie.
A qualitative example is shown in Fig.~\ref{fig:plot_desc_alignment}. Future work will incorporate this global context from movie plots to further improve retrieval performance. 

\vspace{-3em}

\section{Conclusion}
\vspace{-0.5em}

In this work, we introduce a new and challenging \textit{Condensed Movies Dataset} (CMD), containing captioned video clips following succinct and clear storylines in movies. Our dataset consists of long video clips with high level semantic captions, annotated face-tracks, and other movie metadata, and is freely available to the research community. We investigate the task of story-based text retrieval of these clips, and show that modelling past and previous context improves performance. Beside improving retrieval, developing richer models to model longer term temporal context will also allow us to follow the evolution of relationships~\cite{kukleva2020interactions} and higher level semantics in movies, exciting avenues for future work.

\section*{Acknowledgements}
This work is supported by a Google PhD Fellowship, an EPSRC DTA Studentship, and the EPSRC programme grant Seebibyte EP/M013774/1. We are grateful to Samuel Albanie for his help with feature extraction.

\bibliographystyle{splncs}
\bibliography{shortstrings,vgg_local,egbib}
\appendix
\section{Dataset}
\subsection{Character Identity Pipeline} 
\label{sec:app:cidpipeline}
We describe in detail the process of building the character embedding bank mentioned in Sec.~\ref{sec:dataset_collection} of the main paper, and state some figures on the number of annotations obtained. 
We follow a three step \textit{scalable} pipeline to assign character IDs to each of the face-tracks where possible, crucially without any human annotation. First, we use the cast lists obtained for each of the featured movies from IMDb to get a total list of 28,379 actor names. Note we use the names of the \textit{actors} and not characters (the cast lists provide us with the mapping between the two).  $200$ images are then downloaded from image search engines for each of these names. Faces are detected and face-embeddings extracted for each of the faces in the downloaded images. Second, we automatically remove embeddings corresponding to false positives from each set of downloaded images. We achieve this by clustering each of the face-embeddings in the downloaded images into identity clusters (we use agglomerative clustering \cite{jain1988algorithms} with a cosine distance threshold of 0.76\footnote{\label{note_thresh} value found empirically using cross-validation on a subset of manually annotated samples } - embeddings that have a lower similarity than this threshold are \textit{not} merged into the same cluster). We make the assumption that the largest cluster of face-embeddings corresponds to the actor ID that was searched for. If the largest cluster is smaller than a certain threshold (the value 30 is used\textsuperscript{\ref{note_thresh}}) then we remove the actor ID with the conclusion that too few images were found online (commonly the case for relatively unknown cast/crew members). Finally for the remaining actor IDs, the embeddings in the largest cluster are average pooled and L2 normalised into a single embedding. This process leaves us with 13,671 cast members in the \textit{character embedding bank}.
Facetracks are then annotated using the character embedding bank by assigning a character ID when the cosine similarity score between a facetrack embedding and character embedding is above a certain threshold (we use 0.8 as a conservative threshold to prioritize high precision).

\begin{figure}
    \centering
    \includegraphics[width=0.7\textwidth]{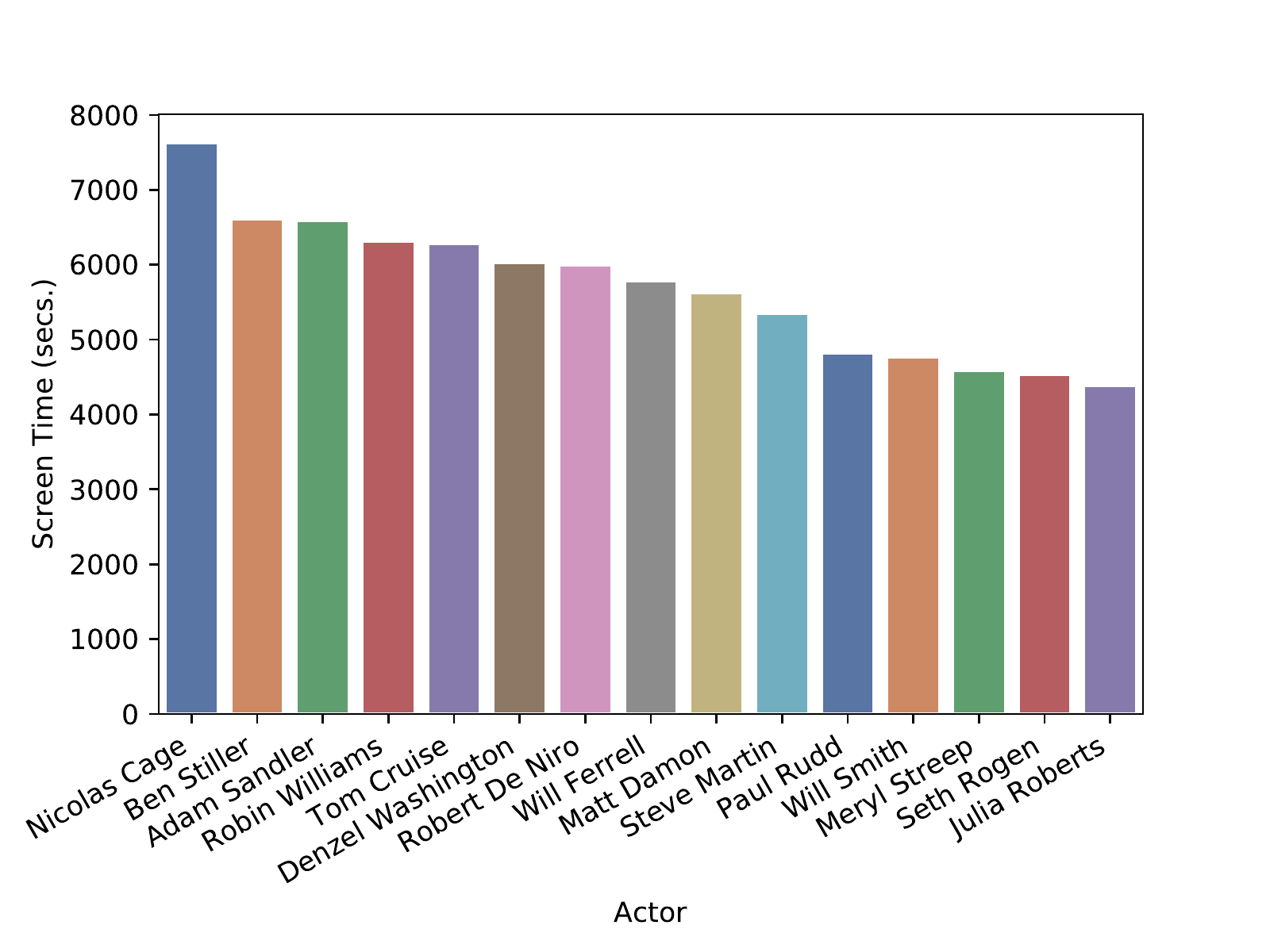}
    \caption{Screen Time of the top 15 most frequent actors recognised by our character identity pipeline, computed as the total duration of face-tracks.}
    \label{fig:actor_screentime}
\end{figure}

The most frequent actors in terms of screen-time automatically labelled by our method can be found in Fig.~\ref{fig:actor_screentime}

\subsection{Aligning Plots to Captions}
\label{sec:app:plotalign}
Aligning video to text has been investigated by a long history of work~\cite{Bojanowski_2015_ICCV}, in particular plot summaries/synopses with films or TV shows\cite{MSA}. These works assume complete and ordered data streams of both video and text, enabling use of the Dynamic Time Warping (DTW) algorithm introduced in~\cite{Sakoe1978DynamicPA}, significantly constraining the problem. The text data for CMD however is Wikipedia plot summaries which do not contain descriptions of every scene in the movie, but rather succinct sentences describing the important events in the film. Further, the video in our case is not the full-length movie but instead key scenes sparsely sampled from the full video. This means that many of the assumptions of DTW do not hold true.

Instead, we assume that each video clip $V$ should be matched with one plot synopsis sentence $S$. Since for our data $|V| < |S|$, some plot sentences are not matched with any video clip, but every video clip does have a matching sentence. This is setting is handled by Jumping Dynamic Time Warping (JDTW)~\cite{jdtw}. We randomly sample 100 movies and manually align the movie clips from CMD to their Wikipedia plot summaries.
\begin{figure}[h]
     \centering
        \includegraphics[width=0.5\textwidth]{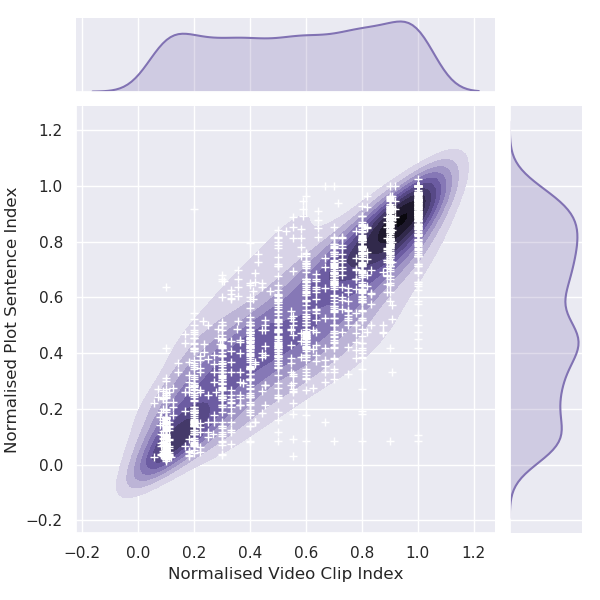}
\caption{\small{\textbf{Plot Coverage: Cumulative distribution of aligned video clips and plot synopsis sentences for 100 randomly sampled movies. Normalised position index indicates their actual index normalised by the total number of video clips / plot synopsis sentences for the movie.}}}
\label{fig:plot_coverage}

\end{figure}
\section{Experiments}
\subsection{Character Module}
\begin{figure}
    \centering
    \includegraphics[width=\textwidth]{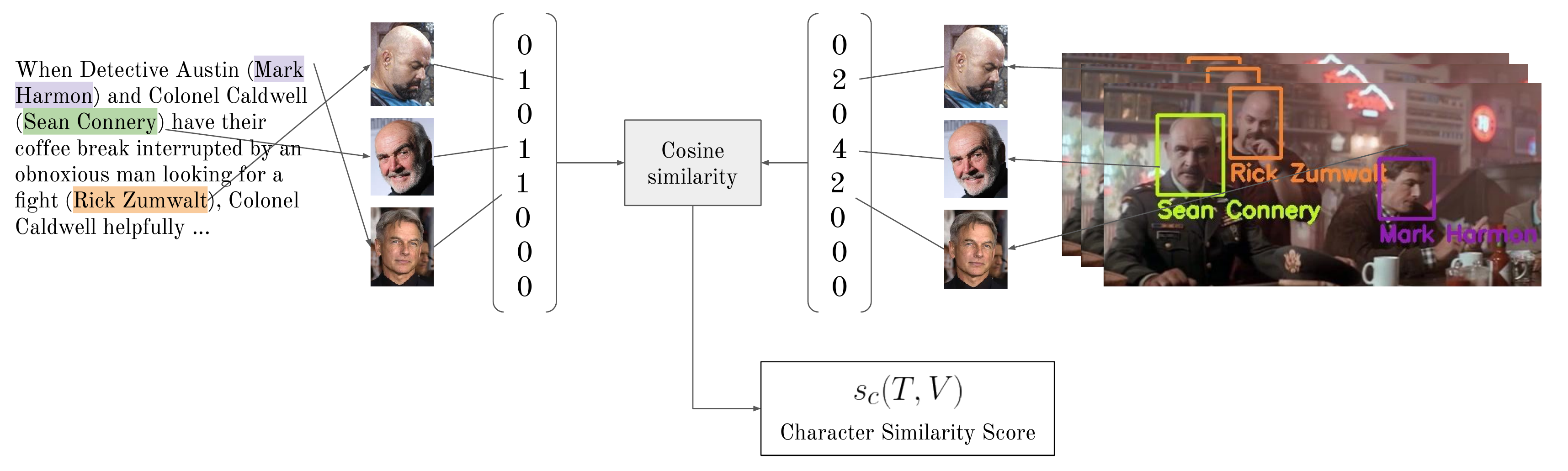}
    \caption{\textbf{Visual Representation of our Character Module.} We show how our character module matches actor names in the caption (left) to actors identified from the video clip (right) using our character embeddings banks. In this example, the video identities are represented by a vector $x$, where each element $x_i$ is the number of facetracks for identity $i$, and the caption identities are represented by a binary vector $y$, where $y_i$ is 1 if identity $i$ is present in the caption and 0 otherwise.}
    \label{fig:char_module}
\end{figure}

The character module, described in Section 4 of the main paper, uses automatically annotated facetracks in the video and actor names in the text to produce a single similarity score. An overview of the character module can be found in Fig.~\ref{fig:char_module}.

\subsection{Context Ablations}
We provide ablations for the best peforming model MoEE + \textit{Contextual Boost Module} (CBM) using a variable number of context videos from the past and future, found in Table~\ref{tab:context-ab}. The results show CBM's general robustness to the amount of context. Past context clips generally outperform  those from the future, which is expected due to the causal nature of the story.

\begin{table}[t]
\caption{\small{Context ablations of the best performing model (MoEE + CBM) on the CMD dataset. Where P$x$ F$y$ denotes $x$ past clips and $y$ future clips used as input to the CBM per target video.}}
\centering
\small
\begin{tabular}{l|ccccc|}
\toprule
 \multicolumn{1}{l|}{} & \multicolumn{5}{c}{Text $\implies$ Video} \\
 \multicolumn{1}{l|}{Context} & \multicolumn{1}{c}{R$@$1} & \multicolumn{1}{c}{R$@$5} & \multicolumn{1}{c}{R$@$10} & \multicolumn{1}{c}{MdR} & \multicolumn{1}{c}{MnR}\\ \midrule
\multicolumn{1}{c|}{P1} & 5.4 & 17.6 & 25.7 & 51 & \multicolumn{1}{r}{260.7}\\
\multicolumn{1}{c|}{P2} & 5.0 & 16.1 & 24.5 & 53 & \multicolumn{1}{r}{250.3} \\
 \multicolumn{1}{c|}{P3} & 5.6 & 17.1 & 25.7 & 50 & \multicolumn{1}{r}{253.8}\\
\multicolumn{1}{c|}{F1} & 4.5 & 15.3 & 23.1 & 58 & \multicolumn{1}{r}{258.7}\\
 \multicolumn{1}{c|}{F2} & 5.1 & 17.0 & 25.5 & 49 & \multicolumn{1}{r}{248.1}\\
\multicolumn{1}{c|}{F3} & 5.4  & 17.1  & 25.9  & 50  & \multicolumn{1}{r}{247.0} \\
 \multicolumn{1}{c|}{P1F1} & 5.0 & 16.4 & 25.3 & 51 & \multicolumn{1}{r}{249.} \\
  \multicolumn{1}{c|}{P3F3} & 5.6 & 17.6 & 26.1 & 50 & \multicolumn{1}{r}{243.9} \\
  \bottomrule 
\end{tabular}\label{tab:context-ab}
\end{table}


\end{document}